\documentclass[a4paper,UKenglish,cleveref, autoref, thm-restate]{lipics-v2021}
\usepackage{graphicx}
\usepackage{amsmath}
\usepackage{makecell}
\usepackage{fancyvrb}
\usepackage{svg}
\usepackage{xcolor}


\title{Automatic Feature Learning for Essence: a Case Study on Car Sequencing}

\author{Alessio Pellegrino}{Dept. of Computer Science and Engineering, University of Bologna, Italy}{alessio.pellegrino@studio.unibo.it}{}{}

\author{Özgür Akgün}{School of Computer Science, University of St Andrews, Scotland \and \url{https://www.st-andrews.ac.uk/computer-science/people/oa86/}}{ozgur.akgun@st-andrews.ac.uk}{https://orcid.org/0000-0001-9519-938X}{}

\author{Nguyen Dang}{School of Computer Science, University of St Andrews, Scotland \and \url{https://www.st-andrews.ac.uk/computer-science/people/nttd/}}{nttd@st-andrews.ac.uk}{https://orcid.org/0000-0002-2693-6953}{}

\author{Zeynep Kiziltan}{Dept. of Computer Science and Engineering, University of Bologna, Italy}{zeynep.kiziltan@unibo.it}{}{}

\author{Ian Miguel}{School of Computer Science, University of St Andrews, Scotland \and \url{https://www.st-andrews.ac.uk/computer-science/people/ijm/}}{ijm@st-andrews.ac.uk}{https://orcid.org/0000-0002-6930-2686}{}

\authorrunning{A Pellegrino, Ö Akgün, N Dang, Z Kiziltan, I Miguel}
\Copyright{Alessio Pellegrino, Özgür Akgün, Nguyen Dang, Zeynep Kiziltan, Ian Miguel}
\hideLIPIcs

\begin{document}

\maketitle

\begin{abstract}
Constraint modelling languages such as {\sc Essence} offer a means to describe combinatorial problems at a high-level, i.e., without committing to detailed modelling decisions for a particular solver or solving paradigm. Given a problem description written in {\sc Essence}, there are multiple ways to translate it to a low-level constraint model. Choosing the right combination of a low-level constraint model and a target constraint solver can have significant impact on the effectiveness of the solving process. Furthermore, the choice of the best combination of constraint model and solver can be instance-dependent, i.e., there may not exist a single combination that works best for all instances of the same problem. In this paper, we consider the task of building machine learning models to automatically select the best combination for a problem instance.
A critical part of the learning process is to define \emph{instance features}, which serve as input to the selection model.  
Our contribution is automatic learning of instance features directly from the high-level representation of a problem instance using a language model. We evaluate the performance of our approach using the {\sc Essence} modelling language with a case study involving the car sequencing problem.
\end{abstract}

\section{Introduction}

In many domains, it has long been observed that there is no single algorithm that performs best on all problems or even on all instances of the same problem ~\cite{rice1976algorithm,kotthoff2016algorithm,kerschke2019automated}. To solve difficult computational problems effectively, it is often beneficial to utilise a \emph{portfolio of algorithms} with complementary strengths. This gives rise to the field of Automated Algorithm Selection (AAS), where the aim is to automatically select the best algorithm(s) from an algorithm portfolio for a given problem instance. Over the last few decades, AAS has been shown to be very successful in various applications across a wide range of domains, including Boolean Satisfiability (SAT)\cite{xu2008satzilla}, Constraint Programming (CP)~\cite{o2008using,liu2021sunny}, AI planning~\cite{vallati2014asap}, and combinatorial optimisation~\cite{kotthoff2015improving}. 

In the CP domain, an algorithm can be seen as a constraint solver (or a specific parameter configuration of a solver). Several studies have demonstrated complementary strengths of constraint solvers~\cite{dang_et_al:LIPIcs.CP.2022.18,dang2022portfolio} and the advantage of using them in combination in a portfolio setting~\cite{o2008using,amadini2015sunny,amadini2016portfolio}. However, the concept of a CP algorithm can be extended beyond the scope of a constraint solver, which often works on a low-level representation of a problem. Those representations are usually less user-friendly and require specific modelling choices to be made about various parts of the problem description. To aid the modelling phase of combinatorial problems, mid-level and high-level constraint modelling languages such as MiniZinc~\cite{nethercote2007minizinc} and {\sc Essence}~\cite{Frisch2008} have been proposed. Accompanying these languages are modelling toolchains that support the automated translation of a mid-level or high-level representation of a problem to the low-level input supported by constraint solvers, such as the MiniZinc Toolchain~\cite{nethercote2007minizinc}, 
{\sc Conjure}~\cite{akgun2022conjure}, 
and {\sc Savile Row}~\cite{nightingale2017automatically}. The translation process involves several modelling and reformulation choices. Making the right combination of modelling and reformulation choices may 
have a significant impact on the performance of the target constraint solver~\cite{akgun2022conjure}. In this context, we can consider an algorithm as a combination of modelling and reformulation configuration and a specific constraint solver. 

Compared with the traditional viewpoint of seeing an algorithm as just a constraint solver, the extended viewpoint 
as a combination of modelling and solver choices can 
result in substantial improvement in the performance of AAS approaches.
However, challenges arise when adopting AAS techniques for this extended context. More concretely, AAS techniques often rely on training Machine Learning (ML) models to predict the best algorithm(s) for a given problem instance based on the instance features. As in any ML application, having a good set of input features is of critical importance. The extracted features must be informative and relevant to not only the given problem instance but also to the performance landscape of the combination of modelling and solver choices on that instance. 

One of the well-known instance features for constraint models are the fzn2feat features~\cite{amadini2014enhanced}. This is a set of 95 features that can be extracted from a representation of a constraint model written in the FlatZinc modelling language~\cite{nethercote2007minizinc}. However, FlatZinc models are low-level representations and can only be obtained after specific decisions on the modelling and reformulation choices have been made. The features extracted are therefore tied to a specific low-level model, which may not be suitable for the task that we aim for, i.e., learning to select among different combinations of low-level models and solvers.

In this work, we propose 
to extract features from the high-level representation of a constraint problem. Instead of having to translate a given problem instance into a low-level representation (i.e., FlatZinc representation) before extracting (fzn2feat) instance features, our approach leverages language models to automatically learn instance features directly from the high-level representation of the problem instance. Compared with the existing fzn2feat feature extraction approach, our approach offers three advantages. First, in contrast to fzn2feat where the features were hand-crafted, our approach learns instance features automatically from the 
textual description of a problem instance. Second, fzn2feat relies on a specific low-level representation of a problem in FlatZinc, while our approach works directly at a high-level representation, which can potentially offer more information for the task of choosing the best combination of models and solvers. Third, as shown empirically, the proposed features, once learnt, are computationally cheaper to extract compared to fzn2feat features. We demonstrate our approach using the {\sc Essence} constraint modelling language via a case study with the car sequencing problem~\cite{gent1998two}.\footnote{\url{https://www.csplib.org/Problems/prob001/}}

In the rest of the paper, after giving the necessary background and discussing the related work in Section \ref{sec:bck}, we introduce in Section \ref{sec:methodology} our approach to AAS and in Section \ref{sec:casestudy} our case study. Then we present in Section \ref{sec:exp} the experimental evaluation of our approach and finally conclude in Section \ref{sec:conclusions}. 





\section{Background and Related Work}
\label{sec:bck}
\textbf{Constraint Modelling Tools} To facilitate the modelling phase of combinatorial problems in CP, several domain-specific languages have been developed. Notable among these are MiniZinc \cite{nethercote2007minizinc} and {\sc Essence}~\cite{Frisch2008}. {\sc Essence} is a high-level language designed to abstract problem modelling using a blend of natural language and discrete mathematics. This abstraction addresses the challenging nature of problem modelling, which demands 
expertise and domain-specific knowledge. {\sc Conjure}~\cite{akgun2022conjure}, a tool designed for {\sc Essence}, incrementally refines an initial {\sc Essence} model into {\sc Essence Prime}, a lower level solver-independent constraint modelling language \cite{nightingale2017automatically}, through a series of transformations. 
Non-trivial transformations may yield multiple effective refinements, resulting in a portfolio of models with varying performances depending on the specific instance and solver used. This creates a complex landscape for selecting the optimal algorithm ({\sc Essence Prime} model and solver combination).

\textbf{ML for Algorithm Configuration and Selection} Algorithm configuration is a field focused on optimizing the hyperparameters of an algorithm to enhance its performance based on criteria such as speed, memory usage, or accuracy. This 
process is essentially a search problem within the hyperparameter space, evaluated against a set of training instances \cite{qu2021general}. Complementary to this is the field of algorithm selection, which involves choosing the best-performing algorithm from a portfolio of pre-tuned options to solve a specific problem instance \cite{lindauer2015autofolio}. Both algorithm configuration and selection often leverage ML techniques to inform their decision-making processes. 

ML algorithms like random forests \cite{breiman2001random} and support vector machines 
\cite{suthaharan2016support} are particularly effective at identifying patterns in input features to predict optimal output, making them well-suited for these tasks. An ML algorithm takes as input a set of data points represented by a set of input features and their corresponding desired output (dataset). The initial dataset is analyzed by the algorithm that produces an ML model designed to address the desired task with a certain degree of correctness in the output. Essentially, an ML model is a function approximation from the feature input space to the desired output space. The efficiency of ML models 
in algorithm selection has been demonstrated in numerous applications~\cite{lindauer2015autofolio, xu2008satzilla}. 

\textbf{Neural Networks and Language Models}
Neural Networks (NNs) represent a powerful paradigm within ML, renowned for their ability to learn complex patterns from large datasets. They are particularly adept at generating features from textual input data \cite{devlin2018bert}, which simplifies the creation of ML models. Since the introduction of AlexNet in 2012 \cite{krizhevsky2012imagenet}, NNs have been successfully applied to a wide array of tasks, such as image classification \cite{simonyan2014very}, text classification \cite{vaswani2017attention}, robotics \cite{browne2003convolutional}, and environmental science \cite{maier2001neural}.

\textbf{Related Work.}
Many AAS tools have been proposed to tackle CSPs. Most notably, SUNNY~\cite{liu2021sunny} and CPHydra~\cite{bridge2012case} use a k-NN approach to compute a schedule of solvers which maximizes the chances of solving an instance within a given timeout, while Proteus~\cite{hurley2014proteus} is a hierarchical portfolio-based approach to CSP solving that does not rely purely on CP solvers: it may choose a SAT solver along with an accommodating CSP-to-SAT translation to solve an instance. Moreover, AAS tools designed for SAT problems can be easily adapted to tackle CSPs (and vice-versa). An empirical evaluation of different AAS approaches for solving CSPs (including SAT portfolios) can be found in~\cite{amadini2013empirical} and~\cite{amadini2014sunny}, which show empirical comparisons between SUNNY and AAS approaches originally proposed for SAT scenarios, such as 3S~\cite{kadioglu2011algorithm} and SATzilla~\cite{xu2008satzilla}.

Language models have previously been applied in CP to generate models from natural language problem descriptions~\cite{tsouros2023holy,almonacid2023towards}. NNs have been used to learn features from the raw trajectories of search algorithms for selecting heuristic algorithms in bin packing problems~\cite{alissa2023automated}. Most relevantly, they have been employed to learn instance features for specific problems, such as the Traveling Salesman Problem (TSP), using transformer architectures~\cite{seiler2023using}. In contrast, our contribution is designed to extract instance features from any {\sc Essence} problem specification.



\section{Methodology}
\label{sec:methodology}
Recall that given a problem class instance written in {\sc Essence} and a set of constraint solvers, we can generate a \emph{portfolio of algorithms} for the instance, where each algorithm is a combination of an {\sc Essence Prime} model and a solver. The aim of our AAS task is to build a prediction model to select from the portfolio the best algorithm (with shortest runtime) for the instance. This task involves two key steps: (i) learning features representing a given {\sc Essence} instance from its raw text content
; and (ii) using the learnt features to predict the best algorithm. 

To address the first step, we propose to employ a Neural Network (NN) that encapsulates a language model to deal with text input. This approach has many advantages. First, language models like Bert have been proven effective in capturing high-level language features \cite{devlin2018bert}, eliminating the need to run a solver to extract the necessary features. Second, NN models can automatically generate the necessary features by starting from the raw input. This eliminates the need for handcrafting an effective feature set.

\begin{figure}[t!]
    \centering
    \includegraphics[width=1\textwidth]{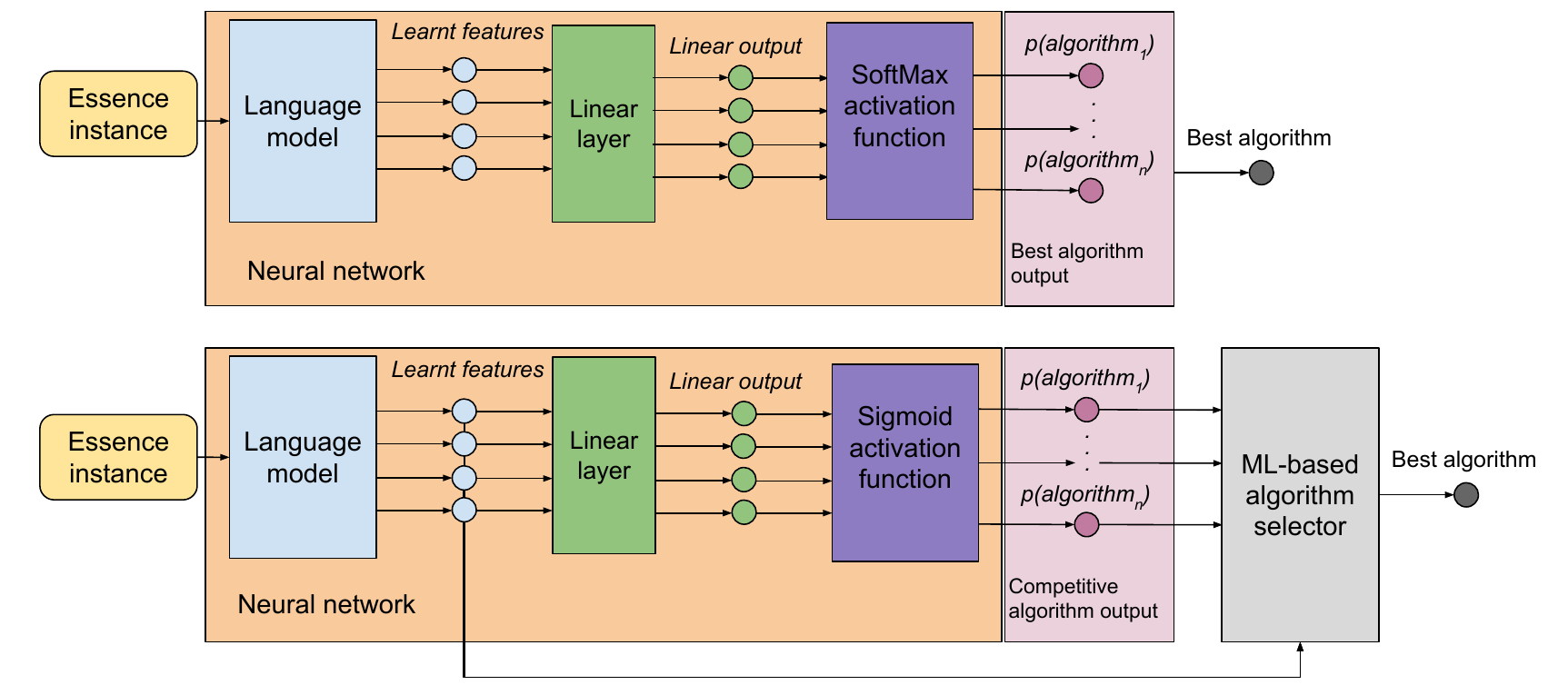}
    \caption{Two possible algorithm selection approaches: entirely NN-based (top) versus hybrid of an NN and an ML-based algorithm selector  (bottom). }
    \label{fig:model}
\end{figure}

For the second step, we consider different options. A possibility is to combine the two steps and address the entire AAS task using a single NN. In this case, the probability associated to an algorithm by the NN indicates its likelihood of being the best and thus the one with the highest probability is deemed as the best. Another possibility is to detach the second step from the first and adopt an ML-based algorithm selector. This gives flexibility in the algorithm selection method, allowing us to leverage state-of-the-art tools as well as to experiment with others. In this case, the probability associated with an algorithm indicates its likelihood to be \textit{competitive}
(that exhibits good performance on the given instance).  
The algorithms along with the produced features are then given as candidates to the algorithm selector which then decides the best one.  Both approaches are depicted in Figure \ref{fig:model}, the details of which are explained in the following subsections.

\subsection{Feature Learning Using a Language Model}
We adopt a language model, a particular NN architecture, to learn a set of features that will be later used to select an algorithm in both options mentioned previously.  The input of such a model is the raw text of the {\sc Essence} instance in tokenized form (where each input word and symbol are transformed into a number), and the output is a feature vector that describes the semantic meaning of the input. In particular, we use 
an 8-bit-quantized \cite{yang2019quantization} version of Longformer\cite{beltagy2020longformer}.\footnote{https://huggingface.co/tororoin/longformer-8bitadam-2048-main} This is a Bert-like \cite{devlin2018bert} architecture whose main advantage is the larger input size (2048 tokens instead of 512) and it has been proven competitive for a fine-tuning task 
\cite{khandelwal2021fine}. In addition to the language model, the NN encapsulates a linear layer to process the features and produce an output. The linear layer comprises a neuron for each of the possible algorithms to choose from. Each neuron receives as input the feature vector produced by the language model. Then it computes the dot product with the learnt weights and adds a bias. The final result is a floating point value for each neuron.

The main difference between the network of the first method (entirely NN-based) and the second one (hybrid of NN and ML-based algorithm selector) lies in the activation function that can transform the output of the linear layer from floating-point values into probabilities to better interpret the NN output.  In the entirely NN-based approach, we want to learn a probabilistic distribution which has, as the most probable value, the best algorithm to choose. To achieve this output, we use the \textit{softMax} activation function that transforms the input sequence into a probability distribution. In the hybrid case instead, we train the NN on a multi-label classification task, where the output comprises probabilities for each algorithm, indicating their \emph{competitiveness} fraction. A higher probability suggests that the algorithm is less likely to be competitive. We consider an algorithm to be competitive if it solves an instance in less than ten seconds or in less than double the time taken by the best-performing algorithm for that instance. For example, if the best algorithm takes 15 seconds, any algorithm that completes the task in under 30 seconds is deemed competitive. To obtain such output, we use the \textit{sigmoid} activation function which transforms each input value to a proper fraction, depending on its magnitude.

\subsection{Algorithm Selection Using the Learnt Features}
\label{sec:aas_using_learnt_features}

Once the NN is trained, the best algorithm for a given {\sc Essence} instance is chosen based on the probabilistic NN output. In the entirely NN-based approach, it is the one with the highest probability. In the hybrid approach,  the probabilistic NN output is fed as input to an ML-based algorithm selector.

As an algorithm selector, 
we can rely on well-known methods such as  Autofolio~\cite{lindauer2015autofolio} and K-means clustering~\cite{ahmed2020k}. The first is a state-of-the-art tool that tunes the underlying model and its hyperparameters to optimize the performance. It can be used both for classification and regression tasks. The second is a clustering algorithm that assigns a cluster to a new instance. As features, these methods can exploit both the language model output and the probabilistic NN output. The features derived from the language model would be useful because they are trained on a similar task, capturing the general semantic structure of the instance. Whereas, the linear layer output 
indicates which algorithms are most likely to perform competitively. By combining the two, the features can encapsulate
both a broad semantic representation of the instance and a specific prediction of the algorithms most likely to be competitive. To combine the features, the two outputs are concatenated, resulting in a vector of floating-point values for the given instance.

To obtain an algorithm selector from K-means clustering, each cluster is associated with the algorithm that resulted the best for the subset of the instances composing the cluster. At prediction time, a new instance is assigned to a cluster and the respective algorithm is selected.

As an alternative ML-based algorithm selector, we can use the probabilistic NN output as an initial filtering mechanism to eliminate the algorithms that are less competitive for a given instance, for instance those with probability less than 0.5. After the filtering, we can rank the remaining candidates based on a certain criterion (measured on the training set) and select the first ranked as the best algorithm.  Possible criteria could be the overall performance or the number of instances where the algorithm wins.

\section{A Case Study with the Car Sequencing Problem}
\label{sec:casestudy}

We evaluate the performance of our approach to AAS using the {\sc Essence} modelling language with a case study involving the car sequencing problem. In this section, we describe the case study. We start with the problem description in {\sc Essence} and the instance set employed in the evaluation. We then present the combinations of (low-level) {\sc Essence Prime} models produced by {\sc Conjure} and constraint solvers, giving rise to a portfolio of algorithms to choose from. Finally, we describe how we obtain a dataset starting from the instance set and the algorithms, and discuss its suitability for an AAS task.  

\subsection{Problem Description and Instance Set}

A series of cars are scheduled for production, each varying due to the availability of different optional features. The assembly line consists of various stations that install these options, such as air conditioning and sunroofs. Each station is designed to handle only a specific percentage of the cars passing through. To ensure that the workload at each station remains manageable, cars requiring the same option must be distributed evenly along the assembly line; clustering of these cars must be avoided to prevent overwhelming any single station. Therefore, cars must be sequenced so that the capacity of each station is not exceeded. For example, if a particular station can only manage a maximum of 50\% of the cars passing through, the sequence must ensure that at most one car in every two requires that option. This sequencing problem is known to be NP-complete \cite{gent1998two}. An {\sc Essence} model for this problem is shown in~\Cref{fig:car_sequencing_essence}.

\begin{figure}[t!]
\begin{verbatim}
given n_cars, n_classes, n_options : int(1..)
letting Slots  be domain int(1..n_cars),
        Class  be domain int(1..n_classes),
        Option be domain int(1..n_options)
given quantity : function (total) Class  --> int(1..),
      maxcars  : function (total) Option --> int(1..),
      blksize  : function (total) Option --> int(1..),
      usage    : relation of ( Class * Option )
find car : function (total) Slots --> Class
such that forAll c : Class . |preImage(car,c)| = quantity(c)
such that forAll opt : Option .
            forAll s : int(1..n_cars+1-blksize(opt)) .
              (sum i : int(s..s+blksize(opt)-1) . 
                toInt(usage(car(i),opt))) <= maxcars(opt)
\end{verbatim}
\caption{Essence model of the car sequencing problem.
}
\label{fig:car_sequencing_essence}
\end{figure}

The {\sc Essence} model defines three integer parameters \textit{n\_cars}, \textit{n\_classes}, and \textit{n\_options} representing the number of cars, classes of cars, and options available, respectively. Using these, three integer domains are defined: \textit{Slots}, \textit{Class}, and \textit{Option}. These domains are used when defining further parameters and decision variables in the model as well as in constraint expressions. Three parameters with function domains are defined to represent the \textit{quantity} of each class of car required, a maximum number of cars (\textit{maxcars}) that can appear in any block of cars, and block size (\textit{blksize}) for each option. The \textit{usage} parameter is a relation that indicates which classes use which options.

The only decision variable (\textit{car}) in the model is a mapping from car production slots to classes. The problem constraints are captured in two top-level constraints (denoted by the keywords \textit{such that}). The first set of constraints ensures that the number of cars in each class matches the required quantity. The second set of constraints ensures that for each option, in any block of \textit{blksize(opt)} consecutive cars, the number of cars requiring that option does not exceed \textit{maxcars(opt)}.

For all experiments in this work, we make use of a large instance set from a previous work~\cite{spracklen2023automated}. It is composed of 10,214 instances, generated using an automated instance generation tool AutoIG~\cite{dang2022framework} for constraint problems, and is publicly available in the {\sc Essence} Catalogue~\cite{EssenceCatalog}. 

\subsection{Combinations of Models and Solvers}

Our algorithm portfolio contains three alternative {\sc Essence Prime} models and four state-of-the-art solvers.  
The solvers are Kissat, Chuffed, CPLEX, and OR-Tools CP-SAT, each chosen for their potential complementary 
characteristics in combinatorial optimization. 
Kissat~\cite{kissat} is a modern clause-learning Satisfiability (SAT) solver.
Chuffed~\cite{Chuffed} is a Constraint Programming (CP) solver enhanced with clause learning.  CPLEX~\cite{IBM_CPLEX} is a commercial Mixed-Integer Programming (MIP) solver that excels in solving problems that heavily use arithmetic constraints. OR-Tools CP-SAT~\footnote{\url{https://developers.google.com/optimization/cp/cp_solver}} is a hybrid solver developed by Google that integrates clause learning, CP-style constraint propagation, and MIP solving methods. 

We use {\sc Savile Row}~\cite{nightingale2017automatically} to target these solvers. 
{\sc Savile Row} is a modelling tool that converts problem models written in {\sc Essence Prime} into the input format required by these solvers and optimises the models based on the characteristics of the specific instance being solved.
The {\sc Essence Prime} models are obtained using {\sc Conjure}~\cite{akgun2022conjure} in its portfolio mode, with variations arising from different representations 
for the \textit{car} decision variable and the \textit{usage} parameter, as well as the way problem constraints are formulated.

The \textit{car} decision variable has two possible representations. The first is a one-dimensional array indexed by cars, containing decision variables with integer domains, where each entry represents the class selected for that car. The other is a two-dimensional Boolean array, indexed by both cars and classes, where a true value indicates the assignment of a car to a class. The \textit{usage} parameter also has two possible representations: a two-dimensional Boolean array or a set of tuples. The second problem constraint in the {\sc Essence} model that refers to the \textit{usage} parameter is refined with an \textit{element} constraint when the Boolean array 
is chosen, instead with a \textit{table} constraint when the set of tuples 
is chosen.

Using a combination of these model fragments, {\sc Conjure} constructs three distinct {\sc Essence Prime} models. The first model $M_1$ has a one-dimensional array of integer variables for \textit{car} and a set of tuples with a \textit{table} constraint for the \textit{usage} parameter. The second model $M_2$ couples the same one-dimensional array for \textit{car} with a Boolean array for  \textit{usage} and the \textit{element} constraint. The third model $M_3$ uses a two-dimensional Boolean array for \textit{car}, and a set of tuples and the \textit{table} constraint for \textit{usage}.

\subsection{Dataset and Algorithm Complementarity}

The combination of three {\sc Essence Prime} models and four constraint solvers results in a total of $12$ algorithms. 
To perform the AAS task, we create a dataset by running the algorithms on the 10,214 car sequencing instances and record their runtime. 
The runtimes are measured on a computer with an AMD EPYC 7763 CPU, where each algorithm is given one CPU core and one hour of cut-off time per instance. We define the overall performance of an algorithm on a given instance set as the average runtime required to solve all the instances. To account for cases where an algorithm does not produce an answer within the given cut-off time, we adopt the Penalised Average Runtime (PAR10) metric from the AAS literature~\cite{lindauer2015autofolio}, where unsolved instances are penalised as 10 times the cut-off time. AAS techniques aim at \emph{minimising} the PAR10 score. 

\begin{figure}[t!]
    \centering
    \includegraphics[width=1\textwidth]{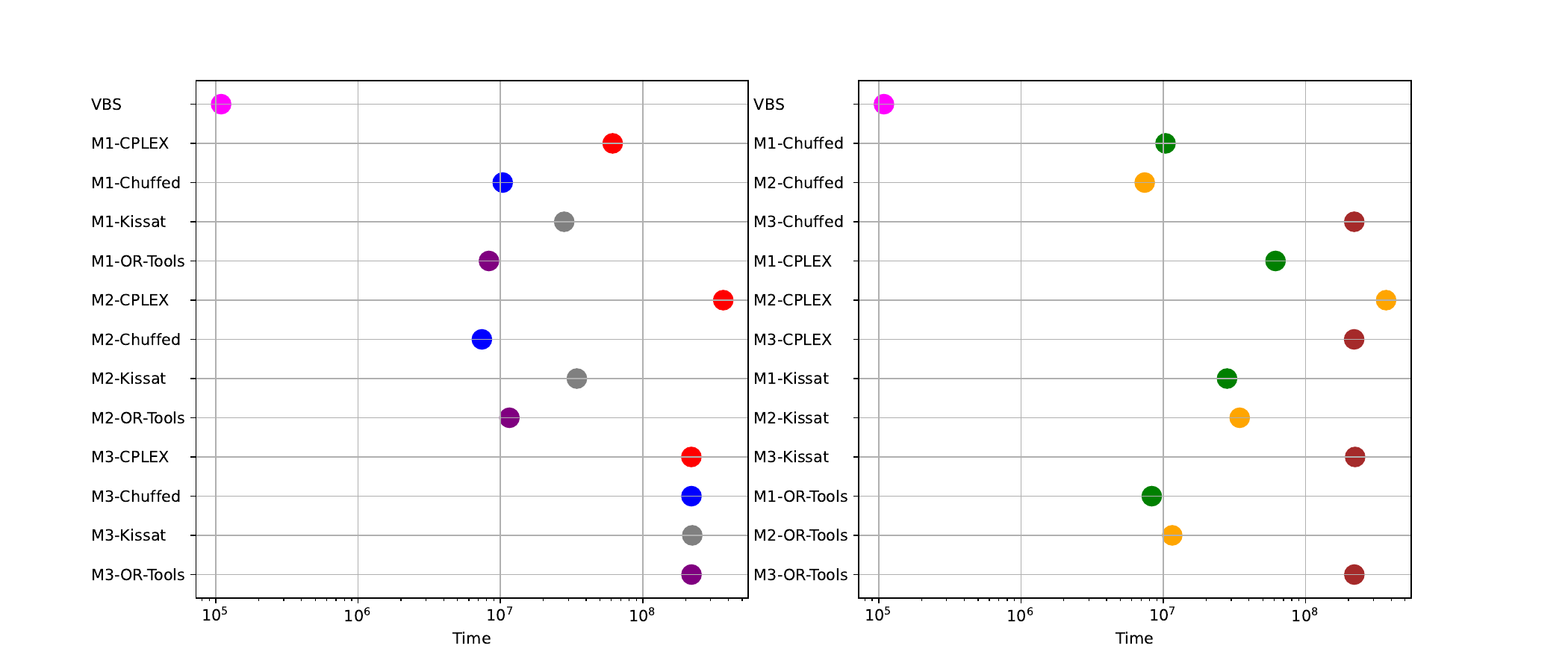}
    \caption{PAR10 value of each algorithm and the VBS on the instance set (lower is better), where the algorithms are grouped by their models (left) or solvers (right).}
    \label{fig:algorithm_time}
\end{figure}

To establish the potential of AAS in this case study, we analyze the performance of each algorithm on the instance set. Figure \ref{fig:algorithm_time} shows the  PAR10 score of the algorithms as well as the Virtual Best Algorithm (VBS), defined as the (hypothetical) algorithm selector that always correctly chooses the best algorithm for each instance. We see that, there is no model (resp. solver) that alone is always the best or worst independently of the coupled solver (resp. model).    
While $M_2$ is fastest with Chuffed, for $M_1$ it is OR-Tools, and these combinations are the two best algorithms. Even though $M_3$ has a much worse score with all the solvers, it does not take part of the worst algorithm, which is $M_2$-CPLEX. Except for the four algorithms involving $M_3$, they all exhibit different performances. Another observation is the big gap between the VBS and the best overall algorithm ($M_2$-Chuffed). We can therefore conclude that the algorithms have complementary strengths and leveraging them via AAS has high potential in this case study. 

\begin{figure}[t!]
    \centering
    \includegraphics[width=0.47\textwidth]{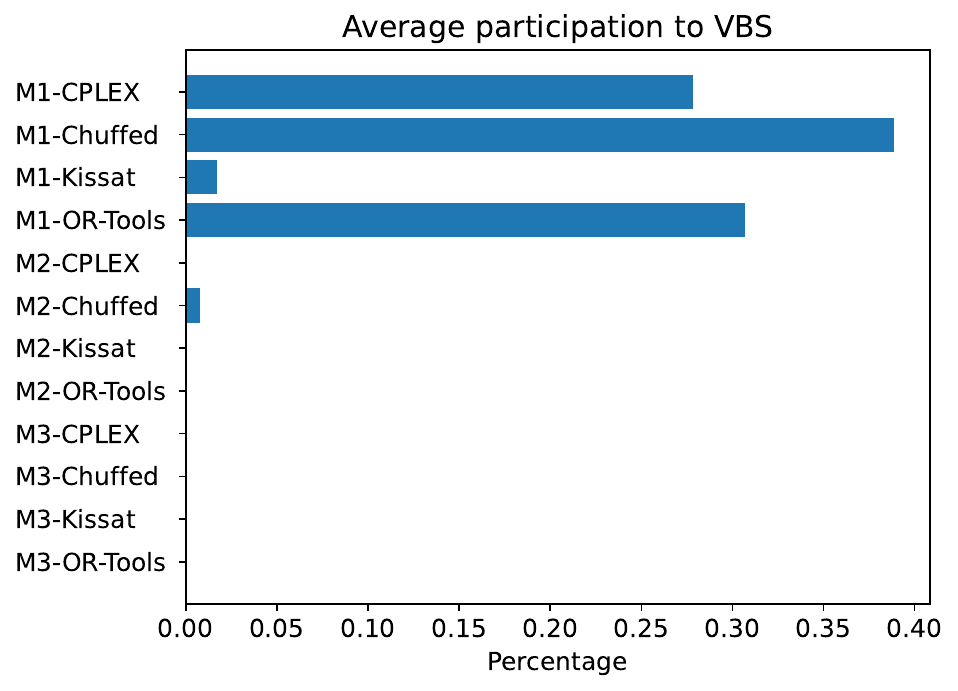}
    \includegraphics[width=0.47\textwidth]{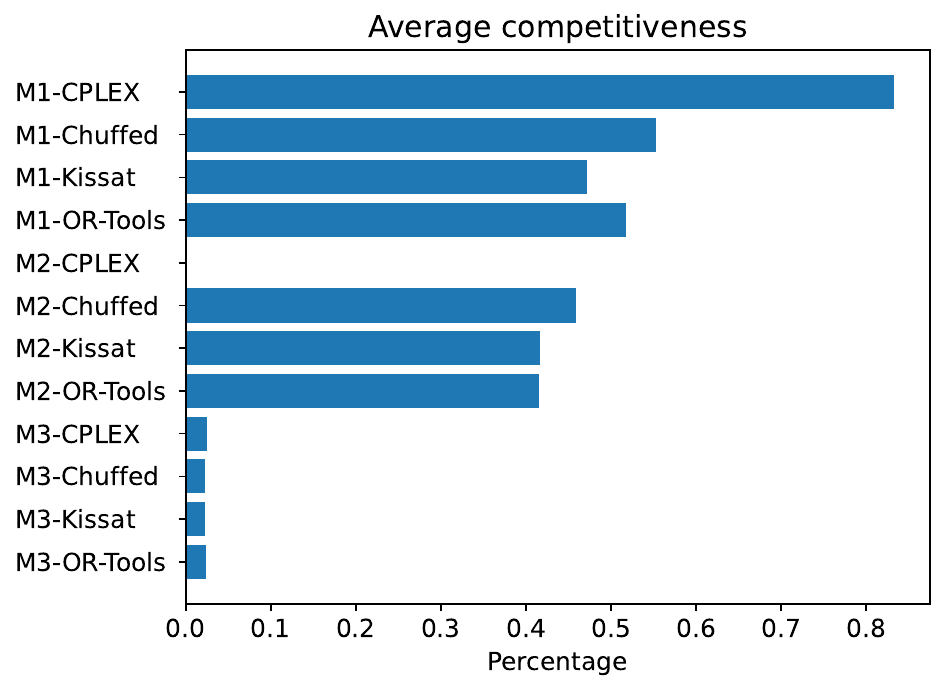}
    \caption{
    Average participation to VBS (left) and average competitiveness (right). 
    }
    \label{fig:algorithm_comp_vb}
\end{figure}
The complementarity of the algorithms in the portfolio can be further observed in Figure ~\ref{fig:algorithm_comp_vb}, where we plot on the left the average participation to VBS (as the percentage of the instances where the algorithm is the best) and on the right the average competitiveness (as the percentage of the instances where the algorithm is competitive). 
We can see that even though $M_2$-Chuffed appears as the best overall algorithm in Figure \ref{fig:algorithm_time}, it is the winner on a fairly small number of instances according to the left plot of Figure \ref{fig:algorithm_comp_vb}. Instead, $M_1$-CPLEX, $M_1$-Chuffed and $M_1$-OR-Tools  have significantly higher numbers of instances where they win. These three algorithms cover a significant part of the instance space. 


While many algorithms do not appear to participate at all to VBS, they are all competitive on some instances (with varying percentages), as shown in the right plot of Figure \ref{fig:algorithm_comp_vb}. An exception is $M_2$-CPLEX which in fact resulted as the worst overall algorithm in Figure \ref{fig:algorithm_time}.  It is typically very difficult for an AAS method to always select the best algorithm for a given instance. At the same, this may not always be necessary, as competitive algorithms could also do well on the instance. We, therefore, expect that being able to choose a competitive algorithm for an instance increases the potential of AAS in our case study.  Indeed, we will provide experimental evidence in Section \ref{sec:exp} that AAS based on predicting the likelihood of an algorithm to be the best performs worse than predicting the likelihood to be competitive.

\section{Experimental Evaluation}
\label{sec:exp}

Having established the potential gain of AAS in the car sequencing case study, in this section, we experimentally evaluate the effectiveness of our approach. 

The research questions (RQs) that we aim to answer in the evaluation are:
\begin{itemize}
    \item RQ1: Can we learn an effective AAS model when combining feature learning and algorithm selection in a single NN model, or do we need to split the learning into two phases (as depicted in~\Cref{fig:model})? 
    \item RQ2: How do the learnt features perform on the AAS task compared to the existing fzn2feat features?
    \item RQ3: What is the feature extraction cost of the learnt features compared to the existing fzn2feat features?
\end{itemize}
We first describe in~\Cref{sec:neural_network_training} how we trained the NN models and then present our study on each RQ in the subsequent sections. 

The experiments are conducted using Python 3.11 in conjunction with PyTorch\footnote{https://pytorch.org/} and scikit-Learn\footnote{https://scikit-learn.org/stable/index.html} for the NN and the K-means clustering, while Python 3.6 was used with Autofolio. \footnote{https://github.com/automl/AutoFolio/tree/master}
The code is publicly available via the project repository. \footnote{https://github.com/SeppiaBrilla/EFE\_project}

\subsection{Neural Network Training}\label{sec:neural_network_training}

All NN models are trained on a GPU with Nvidia A5000 accelerator. \footnote{https://www.nvidia.com/en-us/design-visualization/rtx-a5000/} We trained each NN using a 10-fold cross-validation technique. At each fold, 10\% of the dataset was used as a test set while the remaining 90\% was split into training (90\%) and validation (10\%). For the approaches where the feature learning and algorithm selection are conducted separately, the same data split is used for the ML-based algorithm selector, therefore, if an instance was in the test set of the NN, it was also in the test set of the ML model that used the extracted features. Each network is trained for 10 epochs. For each fold, it took 57,328 seconds, which is around 15.9 hours, to complete the training of each network.

For the entirely NN-based approach where feature learning and algorithm selection are in a single NN model, the training is done using the typical cross entropy loss function for multi-class classification tasks. For the hybrid approach where the NN output is based on algorithm competitiveness, for the first 3 epochs, we used a learning rate of \(1e^{-4}\) and, as a loss function, a weighted version of the Binary Cross-Entropy (BCE) loss that prioritised recall over precision. The formula of the weighted BCE loss function on each sample is shown in~\Cref{math:loss}, where $n$ is the number of algorithms and $y_i$ and $\hat{y}_i$ are the true and the predicted binary labels, indicating whether algorithm $i$ is competitive or not. The first term in this formula represents the recall metric and is weighted twice over the second term.

\begin{equation}
{\mathcal{L}_{\text{BCE}}(y, \hat{y}) = - \frac{1}{n} \sum_{i=1}^n} \left[ 2y_i \log(\hat{y}_i) + (1 - y_i) \log(1 - \hat{y}_i) \right]
\label{math:loss}
\end{equation}
For the next $6$ epochs, we dropped the custom weights to use the normal BCE loss. The only notable change between epochs 3 to 6 and 6 to 10 was the change of learning rate that was \(1e^{-4}\) for epochs 3 to 6 and \(1e^{-5}\) for the final 4 epochs. For the whole training process, we used stochastic gradient descent as an optimizer for the model.

We leave as future work a more systematic study of which training schedules and hyper-parameter configurations are best suited to our task.  The current decision is based on a small manual tuning study. The intuition behind splitting the training into different phases is as follows. At the first stage of the training process (the first $6$ epochs), we prioritise recall over precision. If an algorithm is not competitive but is predicted as so, it may be incorrectly chosen by the algorithm selector and could potentially result in a larger performance loss (in PAR10 score), therefore, the first term in~\Cref{math:loss} is weighted higher to emphasise it.

\subsection{Feature Learning and Algorithm Selection: Combining vs Splitting}

\begin{figure}[t!]
    \centering
    \includegraphics[width=0.49\textwidth]{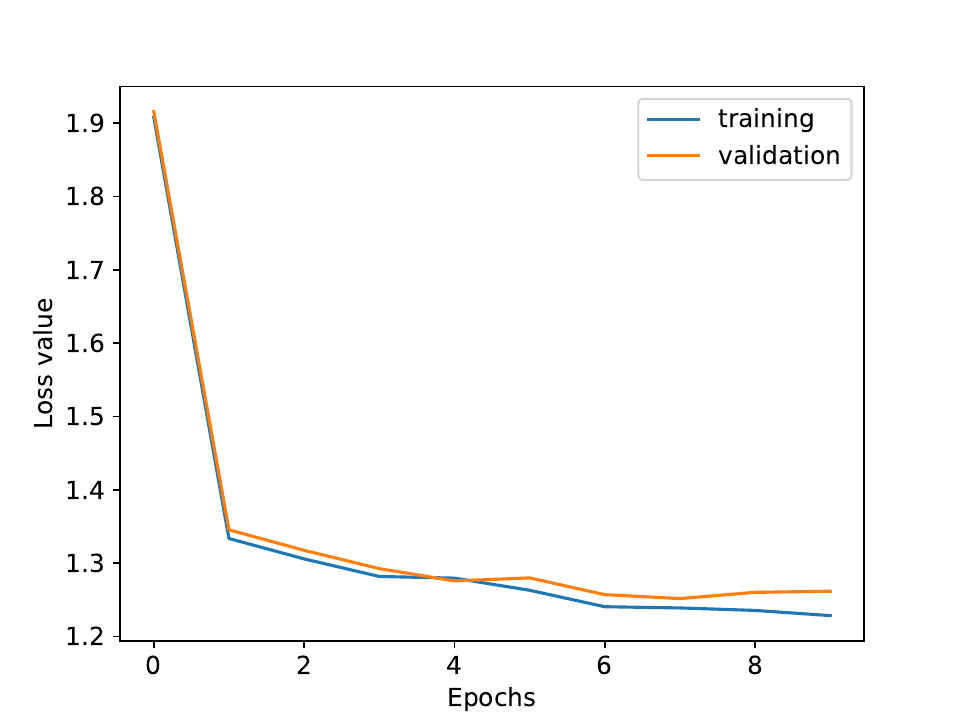}
    \includegraphics[width=0.49\textwidth]{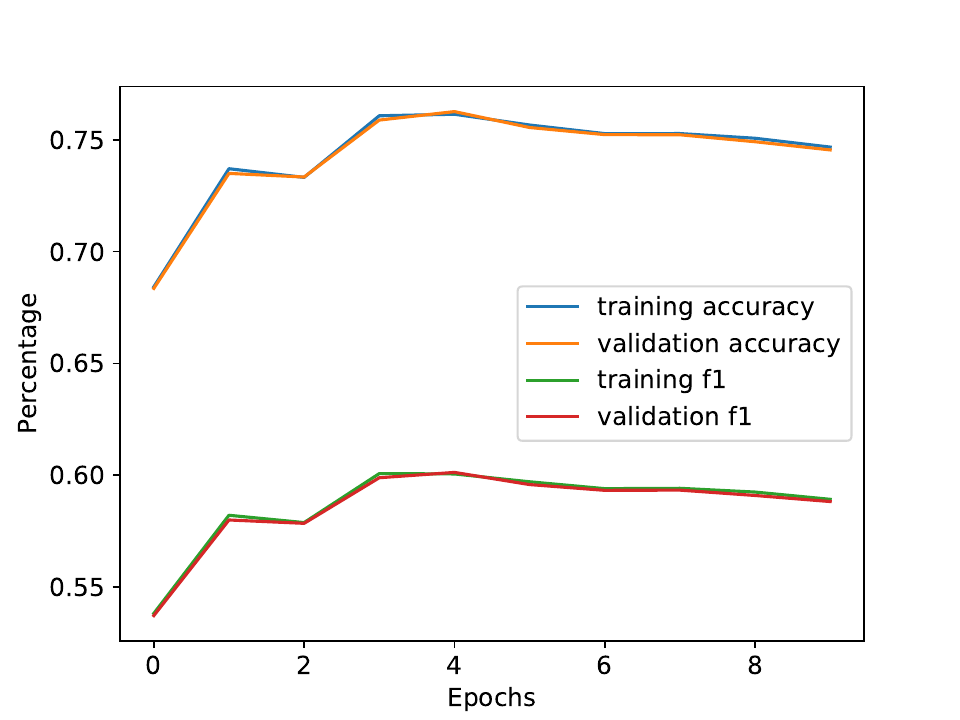}
    \includegraphics[width=0.49\textwidth]{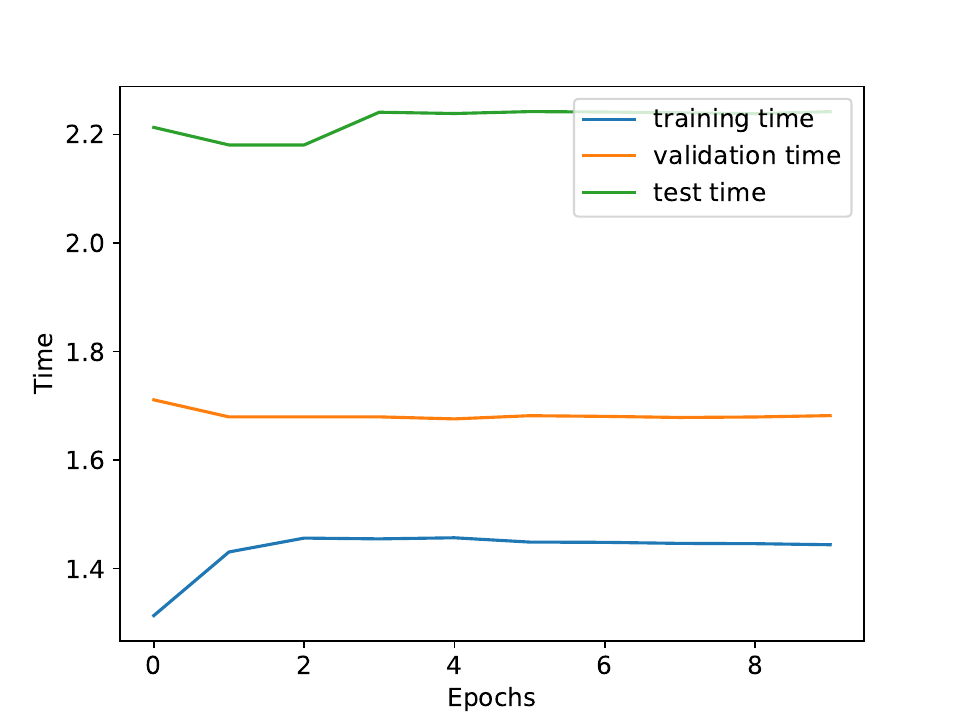}
    \caption{Training progress of the combined learning approach in one fold, shown by the cross entropy loss (top left), accuracy and F1 score (top right), and PAR10 score (bottom). The PAR10 score is normalised into the range $[0,1]$ using $M_2$-Chuffed (the best overall algorithm) and the VBS.}
    \label{fig:loss_fully_neural}
\end{figure}

In this section, we investigate RQ1: Can we learn an effective AAS model when combining both feature learning and algorithm selection in a single NN model, or do we need to split the learning into two phases (as depicted in~\Cref{fig:model})?

\Cref{fig:loss_fully_neural} presents an example of the training progress of the combined learning approach in one fold. Although the cross entropy loss value seems to indicate favourable results, the performance of the learnt network at each epoch in terms of accuracy and F1 score, as well as (normalised) PAR10 score, do not improve after the third epoch. The associated values in both training and the validation sets reach stagnation after that point. We observed the same pattern after having repeated the experiment across multiple folds. This observation highlights the challenges of training a combined learning approach for the AAS task.  

\begin{figure}[t!]
    \centering
    \includegraphics[width=0.49\textwidth]{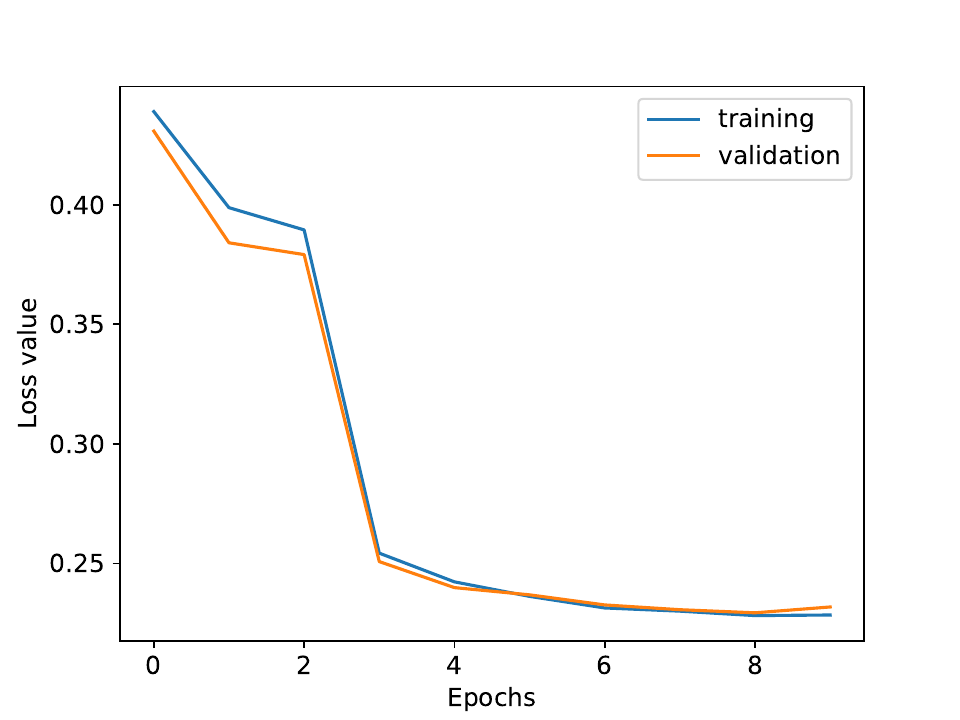}
    \includegraphics[width=0.49\textwidth]{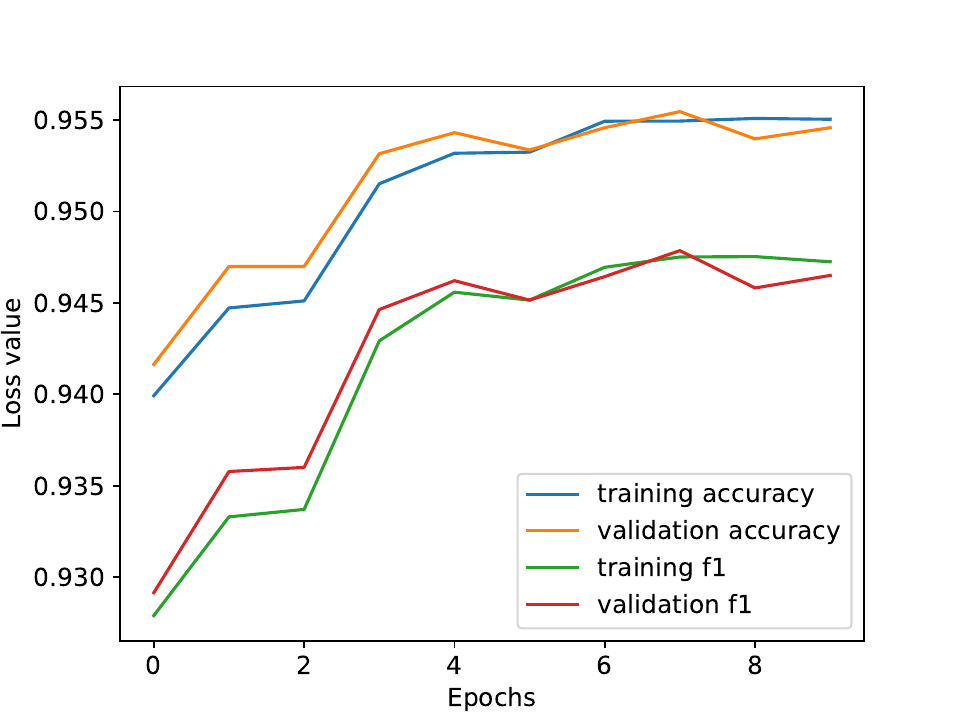}
    \caption{Training progress of the split learning approach in one fold, shown by the cross entropy loss (left), and accuracy and F1 score (right).}
    \label{fig:loss_mixed}
\end{figure} 

One possible explanation for the difficulty of training 
is the fact that when treating the AAS task as a multi-class classification task, the training data is potentially 
highly imbalanced. For instance, some algorithms may win only on a small number of instances, making it difficult to predict them correctly, even though they may have a significant contribution to the overall PAR10 score of the algorithm selector. We mitigate this issue in our split learning approach by replacing the multi-class classification task with a multi-label classification task. Instead of predicting the best algorithm, the output layer of our feature learning network will predict the competitiveness of each algorithm. In fact, this change allows us to train the network more effectively. As illustrated in~\Cref{fig:loss_mixed}, the accuracy and F1 score now improve steadily during the training process. This study indicates that splitting the learning into two separate parts (feature learning and algorithm selection) is more effective. Therefore, we will adopt this approach in the remaining evaluation.

\subsection{Learnt Features vs fzn2feat}

In this section, we investigate RQ2: How do the learnt features perform on the AAS task compared to the existing fzn2feat features?

Our learnt features are the concatenation of the language model output and the probabilistic output of the NN, as illustrated in the bottom part of ~\Cref{fig:model}. As an ML-based algorithm selector, we adopt AutoFolio~\cite{lindauer2015autofolio} and K-means clustering~\cite{ahmed2020k}, as mentioned in ~\Cref{sec:aas_using_learnt_features}. With these selectors, we can use either our NN-based or the fzn2feat features. We refer to the four possible combinations as NN-Autofolio, fzn2feat-Autofolio, NN-Kmeans, and fzn2feat-Kmeans.

In addition to the algorithm selectors named above, our feature learning method offers other possibilities for algorithm selection. As a by-product of the feature learning process, we have a prediction model that tells us which algorithms are less competitive (with probability less than 0.5) for a given instance. As described in~\Cref{sec:aas_using_learnt_features}, this information can be used to filter out the less-promising algorithms for that particular instance. Among the remaining ones, we can select the best algorithm based on a specific criterion (measured on the training set), such as the PAR10 score or the number of instances where the algorithm wins. We refer to these simple selection approaches as NN-based Single Best Selection (NN-SBS) and Winner Selection (NN-WS), respectively.

\begin{figure}[t!]
    \centering
    \includegraphics[width=0.49\textwidth]{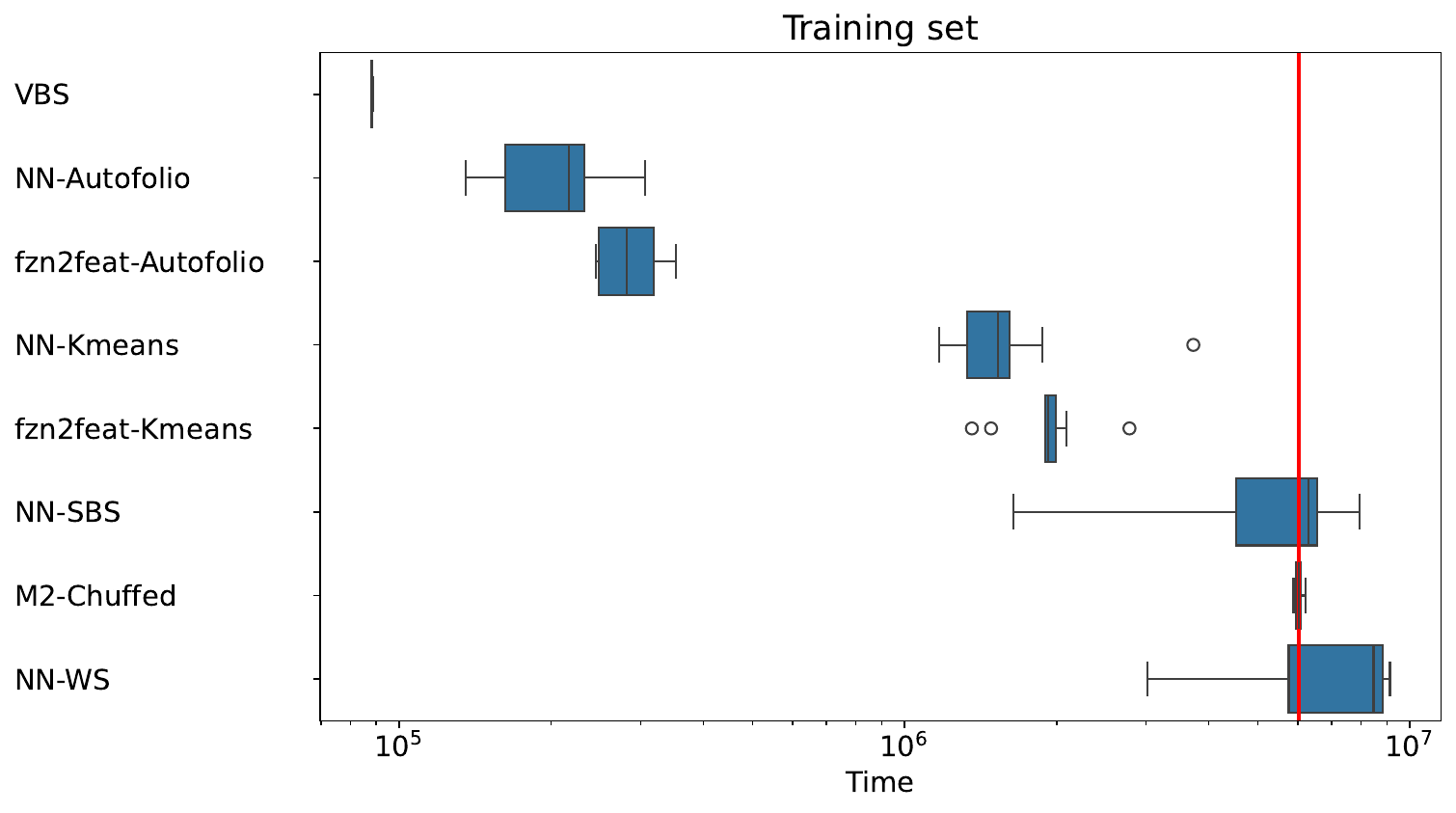}
    \includegraphics[width=0.49\textwidth]{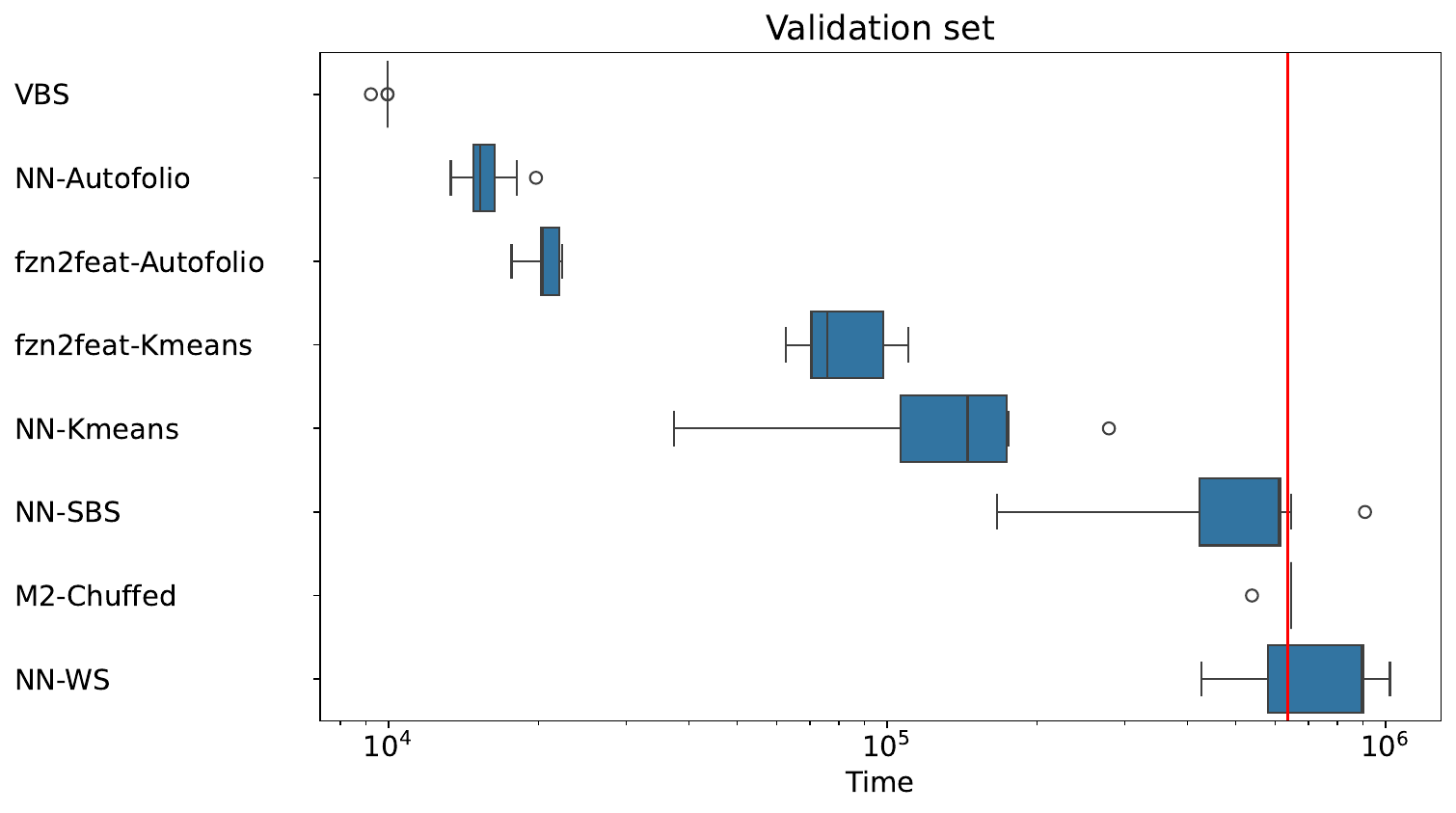}
    \includegraphics[width=0.49\textwidth]{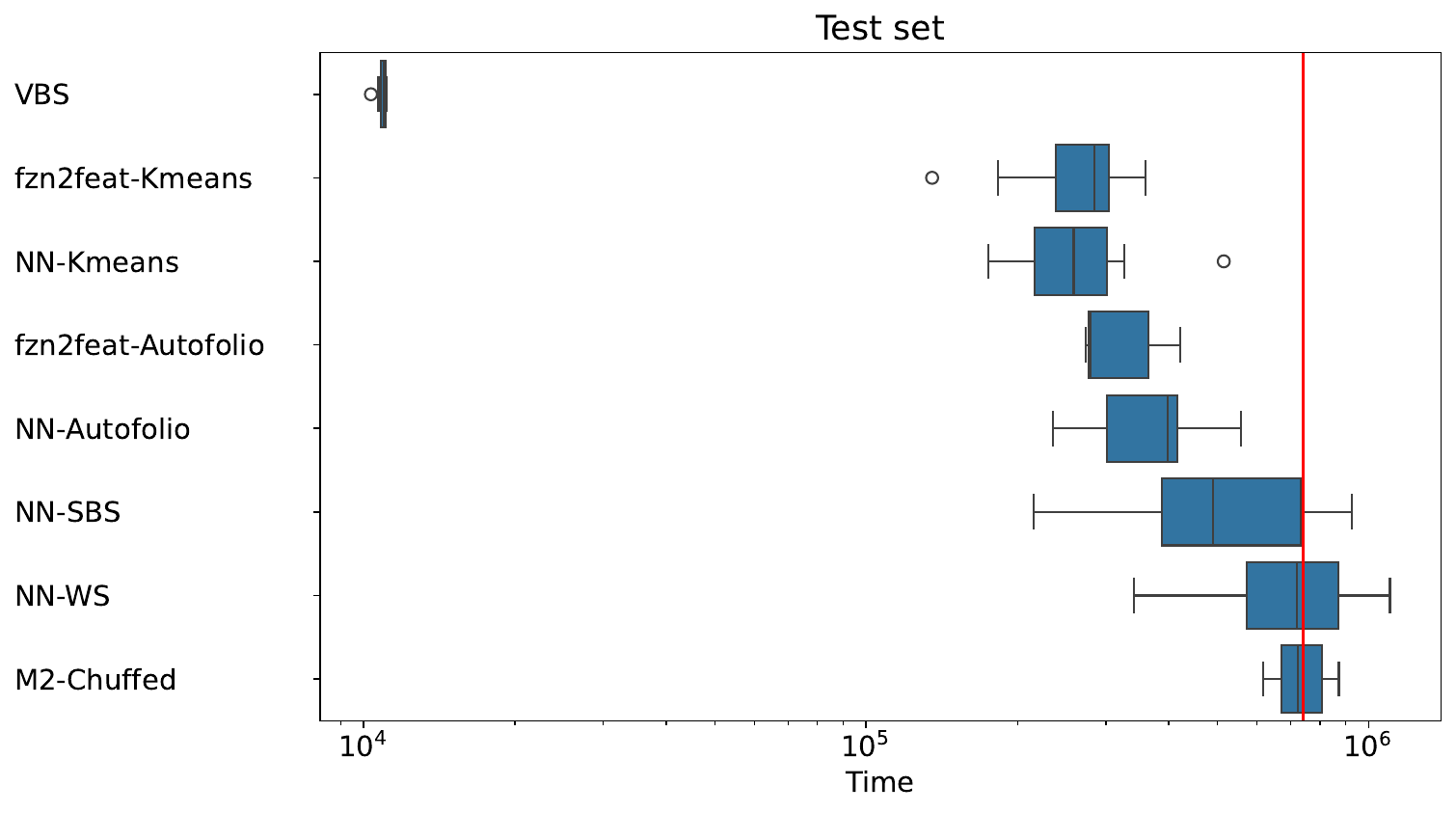}
    \caption{PAR10 scores of different AAS approaches across 10 folds. $M_2$-Chuffed is the best overall algorithm in the portfolio and its mean PAR10 score is shown with the red line.  Reported prediction time includes the feature computation time.}
    \label{fig:final_res}
\end{figure}

\Cref{fig:final_res} presents the PAR10 scores of all the approaches described above on the training, validation and test sets across 10 folds. All four approaches using algorithm selectors surpass the performance of $M_2$-Chuffed (the best overall algorithm) and the two other simple selection methods (NN-SBS and NN-WS), confirming the effectiveness of learning AAS models using either feature set. 
Interestingly, AutoFolio offers significantly better performance than K-means on the training and the validation sets, but its generalisation is reduced as K-means is able to close the gap on the test set. 

Compared to fznfeat, our learnt feature set provides competitive performance, which indicates the effectiveness of the NN-based feature learning process. When combined with K-means, our  feature set provides better overall performance on all the training, validation and test sets, although the difference between the two becomes less visible on the test set. When combined with AutoFolio, the fzn2feat methods offer slightly better average performance on the test set, although the learnt features do produce better on some folds. 


AutoFolio is an algorithm selector that incorporates multiple state-of-the-art candidate ML models. It comes with a default selection model and that is what we have adopted in all the experiments so far. This is not necessarily the best choice, as the best model can depend on the specific scenario. AutoFolio includes an option to search in the vast space of several ML models and for their hyper-parameter configuration using the hyper-parameter optimisation tool SMAC~\cite{hutter2011sequential}. To investigate the effectiveness of the two feature sets further, we conducted a new set of experiments where we allowed AutoFolio to be tuned. The tuning is done using SMAC in a 10-fold cross validation fashion. We let SMAC run for a maximum amount of 5 CPU hours on a machine with an AMD EPYC 7763 CPU. 

\begin{figure}[t!]
    \centering
    \includegraphics[width=0.49\textwidth]{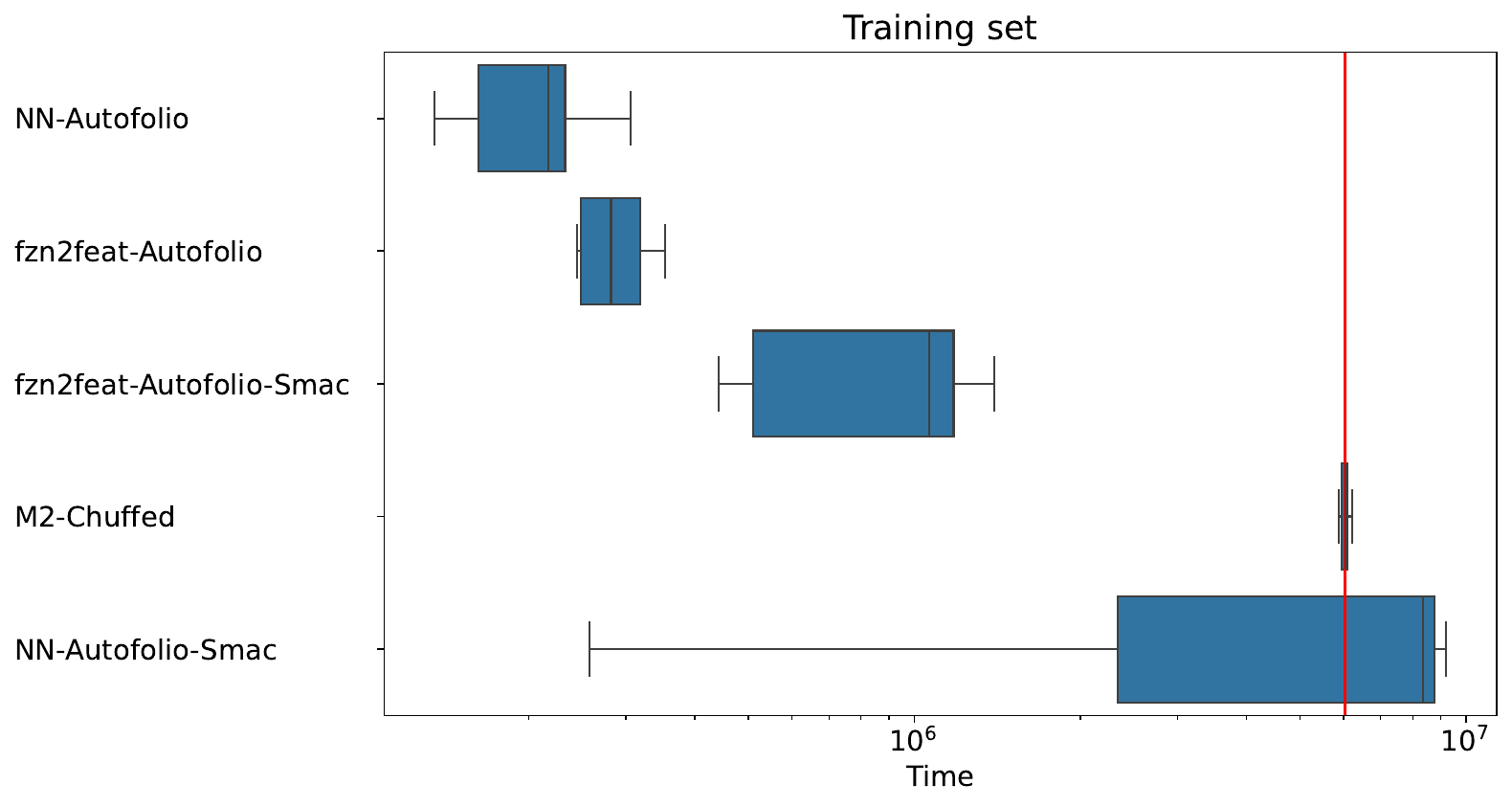}
    \includegraphics[width=0.49\textwidth]{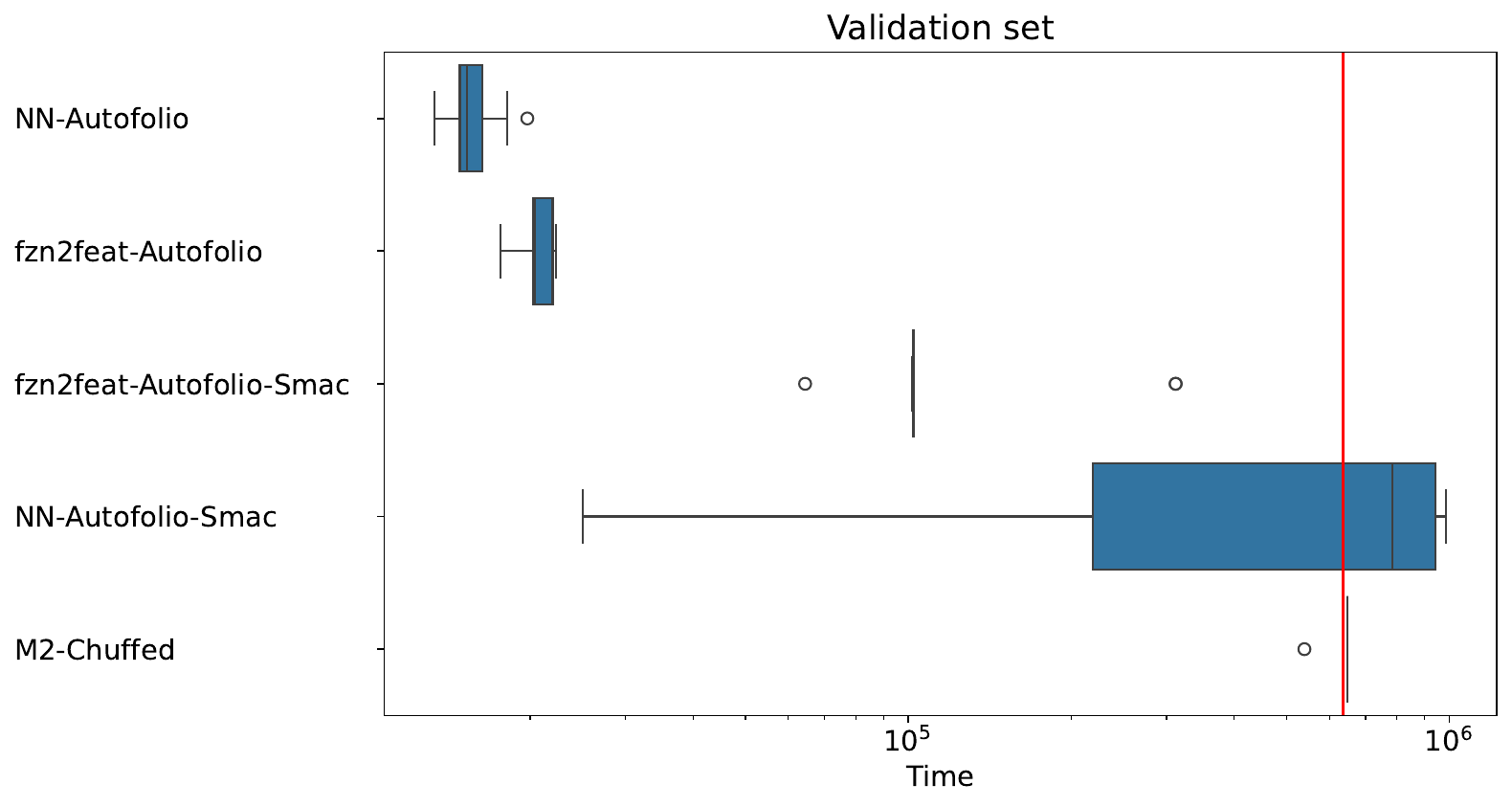}
    \includegraphics[width=0.49\textwidth]{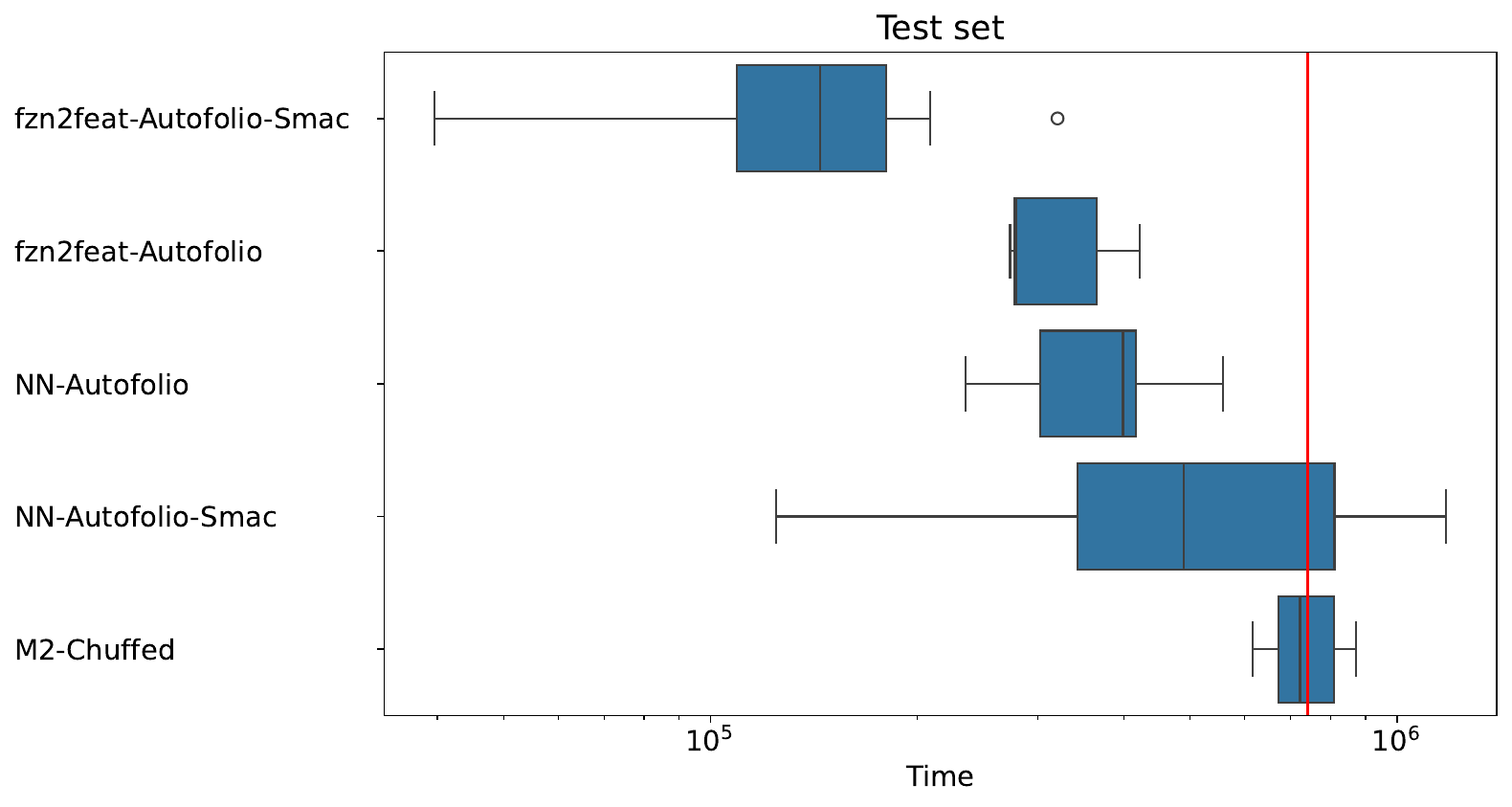}
    \caption{PAR10 scores of Autofolio (tuned with SMAC or not) across 10 folds. $M_2$-Chuffed is the best overall algorithm in the portfolio and its mean PAR10 score is shown with the red line.  Reported prediction time includes the feature computation time.}
    \label{fig:final_res_tune}
\end{figure}

\Cref{fig:final_res_tune} shows the PAR10 scores of AutoFolio coupled with either feature set, with and without tuning. The tuning is very effective when the fzn2feat features are used as input. Surprisingly, when the NN-based features are used, there is a large variance in the performance of the tuned version on all three datasets.
One potential explanation for this observation is that the number of features obtained from NN is very high (783 features) compared to fzn2feat (only 95 features). AutoFolio makes use of classical ML models such as random forests, and those might not be best suited to work on a very high dimensional input space. There are two potential ways to mitigate this issue. First, instead of using AutoFolio, we can try developing an NN-based algorithm selector, which may be better suited to be used with our learnt features. Second, we can try reducing the amount of features produced by the language model by imposing additional linear layers between the language model and the output layer, which may help to compress the learnt feature space. We leave the investigation of these options for future work.



\subsection{Feature Extraction Cost}

In this section, we investigate RQ3: What is the feature extraction cost of the learnt features compared to the existing fzn2feat features?

As indicated in Table \ref{tab:times}, a significant advantage of the NN-based approach is the time required to extract features from an instance. It consistently took less than 0.38 seconds to produce a result, whereas fzn2feat took up to 33 seconds. However, it is important to note that this speed advantage is contingent on the availability of a discrete graphics card, as NNs perform faster on GPUs. 

\begin{table}[t!]
\centering
\begin{tabular}{lllll}
               & Median & Mean & Max   & Min  \\
fzn2feat       & 6.71   & 5.38 & 33.68 & 0.80 \\
NN & 0.02   & 0.02 & 0.38  & 0.02 \\
\end{tabular}
\caption{Statistics to compute a feature vector in seconds across all the instances.}
\label{tab:times}
\end{table}

\section{Conclusions}
\label{sec:conclusions}
In this paper, we explored the use of automatic feature learning for algorithm selection in the context of the car sequencing problem, leveraging the high-level constraint modelling language Essence. Our approach employed a language model to learn instance features directly from the problem descriptions, which were then used to predict the best algorithm for solving each instance. 

Our experiments demonstrated that the learnt features could effectively be utilized within two different algorithm selection strategies (AutoFolio and K-means clustering). Both strategies showed promise, but each had its own strengths and weaknesses. 
The tuning experiments with AutoFolio highlighted the importance of careful feature set selection and tuning, especially given the high dimensionality of the learned features.

Despite these challenges, our results indicate that NN-based feature extraction offers a viable and efficient alternative to traditional methods, with significantly lower computational costs for feature extraction. However, the instability observed in the performance of tuned AutoFolio with NN-based features suggests further refinements are necessary. Future work could involve developing an NN-based algorithm selection approach tailored to handle high-dimensional feature spaces more effectively or incorporating feature compression techniques to enhance stability.

Overall, this study highlights the potential of ML and automatic feature learning in enhancing algorithm selection processes for combinatorial problems, paving the way for more adaptive and efficient solving techniques in various application domains.

\section*{Acknowledgements}
This work was supported by the European Union's Justice programme, under GA No 101087342, POLINE (Principles Of Law In National and European VAT) and by a scholarship from the Department of Computer Science and Engineering of the University of Bologna.

\bibliography{references}
\end{document}